\documentclass{article}

\usepackage{microtype}
\usepackage{graphicx}
\usepackage{booktabs} % for professional tables
\usepackage{subcaption}
\usepackage{tikz}
\usepackage{caption}
\usepackage{pgfplots}
\usepackage{subcaption}
\usepackage{siunitx}
\usepackage{soul}
\usepackage{amssymb}

\usepackage{hyperref}
\usepackage{enumitem}
\usepackage[algo2e,ruled,vlined]{algorithm2e}

\newcommand{\squishlist}{
   \begin{list}{$\bullet$}
    { \setlength{\itemsep}{0pt}      \setlength{\parsep}{3pt}
      \setlength{\topsep}{3pt}       \setlength{\partopsep}{0pt}
      \setlength{\leftmargin}{1.5em} \setlength{\labelwidth}{1em}
      \setlength{\labelsep}{0.5em} } }
      
\newcommand{\squishend}{
    \end{list}  }

\usepackage[accepted]{mlsys2022}

\mlsystitlerunning{Flower: A Friendly Federated Learning Framework}

\begin{document}

\twocolumn[
\mlsystitle{Flower: A Friendly Federated Learning Framework}

% unless you've provided the [accepted] option to the mlsys2022


\mlsyssetsymbol{equal}{*}

\begin{mlsysauthorlist}
\mlsysauthor{Daniel J. Beutel}{cam,adap}
\mlsysauthor{Taner Topal}{cam,adap}
\mlsysauthor{Akhil Mathur}{nbl}
\mlsysauthor{Xinchi Qiu}{cam}
\mlsysauthor{Javier Fernandez-Marques}{ox}
\mlsysauthor{Yan Gao}{cam}
\mlsysauthor{Lorenzo Sani}{unibo}
\mlsysauthor{Kwing Hei Li}{cam}
\mlsysauthor{Titouan Parcollet}{au}
\mlsysauthor{Pedro Porto Buarque de Gusm\~{a}o}{cam}
\mlsysauthor{Nicholas D. Lane}{cam}
\end{mlsysauthorlist}

\mlsysaffiliation{cam}{Department of Computer Science and Technology, University of Cambridge, UK}
\mlsysaffiliation{adap}{Adap, Hamburg, Hamburg, Germany}
\mlsysaffiliation{nbl}{Nokia Bell Labs, Cambridge, UK}
\mlsysaffiliation{au}{Laboratoire Informatique d'Avignon, Avignon Université, France}
\mlsysaffiliation{ox}{Department of Computer Science, University of Oxford, UK}
\mlsysaffiliation{unibo}{Department of Physics and Astronomy, University of Bologna, Italy}

\mlsyscorrespondingauthor{Daniel J. Beutel}{daniel@adap.com}

\mlsyskeywords{Machine Learning, MLSys}

\vskip 0.1in

\begin{abstract}
Federated Learning (FL) has emerged as a promising technique for edge devices to collaboratively learn a shared prediction model, while keeping their training data on the device, thereby decoupling the ability to do machine learning from the need to store the data in the cloud. However, FL is difficult to implement realistically, both in terms of scale and systems heterogeneity. Although there are a number of research frameworks available to simulate FL algorithms, they do not support the study of scalable FL workloads on heterogeneous edge devices. 

%Furthermore, these frameworks are designed to simulate FL in a server environment and hence do not allow experimentation in distributed settings for a large number of clients.
%
%at a realistic scale, considering the heterogeneity in mobile devices, e.g., different programming languages, frameworks, and hardware accelerators.

In this paper, we present Flower -- a comprehensive FL framework that distinguishes itself from existing platforms by offering new facilities to execute large-scale FL experiments, and consider richly heterogeneous FL device scenarios. Our experiments show Flower can perform FL experiments up to \textit{15M in client size} using only a pair of high-end GPUs. Researchers can then seamlessly migrate experiments to real devices to examine other parts of the design space. We believe Flower provides the community a critical new tool for FL study and development. 

% incorporates two new important facilities  

% framework that brings to the research community important new facilities in support of the invention and study of federated algorithms. 

% which is both 

% agnostic towards heterogeneous clients (e.g., mobile and embedded devices) and supports the scaling of evaluations under a large number of clients 

% %also scales to a large number of clients, including mobile and embedded edge devices.
% Flower allows researchers to port existing workloads with little overhead, regardless of the programming language or ML framework used, while also allowing flexibility to experiment with novel approaches to advance the state-of-the-art. We describe the design goals and implementation considerations of Flower and show our experiences in evaluating the performance of FL across clients with heterogeneous computational and communication capabilities.
\end{abstract}
]



\printAffiliationsAndNotice{}  % leave blank if no need to mention equal contribution
%\printAffiliationsAndNotice{\mlsysEqualContribution} % otherwise use the standard text.

\section{Introduction}
\label{introduction}
% TODO Why FL is important (privacy, data movement, legislation[GDPR]); why we need a tool a that has to work for both research and industry and why this tool needs to address scalability, system heterogeneity HW and SW, different ML frameworks, privacy

% TODO Consider the main aspects mentioned before.

% TODO To emphasise that Language-specific frameworks (at least for clients is not ideal) look into: \url{https://www.mdpi.com/1424-8220/21/1/167 }

%\input{mlsys2022style/figure1}

There has been tremendous progress in enabling the execution of deep learning models on mobile and embedded devices to infer user contexts and behaviors~\cite{10.5555/3327144.3327315,DBLP:journals/corr/abs-1906-05721,malekzadeh2019privacypreserving,10.1145/3300061.3345447,10.1145/3319535.3354209,10.1109/ISCA.2016.31,DBLP:conf/mobisys/GeorgievLMC17}.  This has been powered by the increasing computational abilities of mobile devices as well as novel algorithms which apply software optimizations to enable pre-trained cloud-scale models to run on resource-constrained devices. However, when it comes to the training of these mobile-focused models, a working assumption has been that the models will be trained centrally in the cloud, using training data aggregated from several users.

\begin{figure}[!h]
    \small
    \centering
    % \includegraphics[width=0.5\textwidth]{diagrams/framework.pdf}
    % \parjump{}
    %\includegraphics[width=0.4\textwidth]{mlsys2022style/diagrams/fig1_colour.pdf}
    \includegraphics[width=0.43\textwidth]{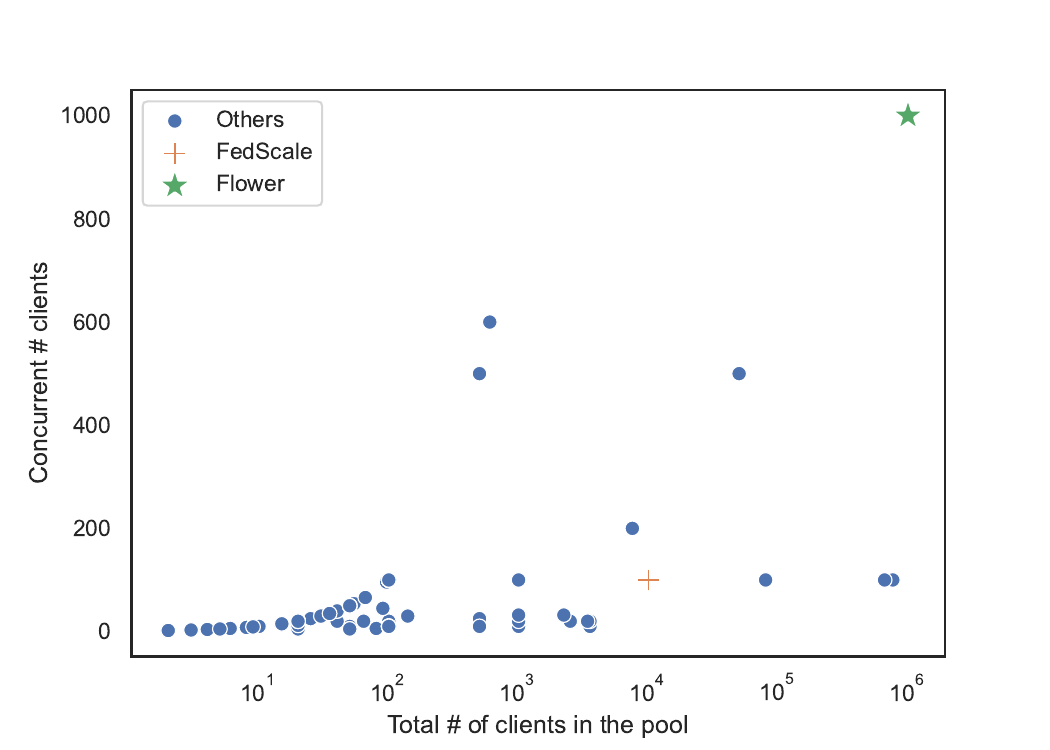}
    \vspace{-0.3cm}
    \captionsetup{font=small,labelfont=bf}
    \caption{Survey of the number of FL clients used in FL research papers in the last two years. Scatter plot of number of concurrent clients participated in each communication round (y-axis) and total number of clients in the client pool (x-axis). The x-axis is converted to log scale to reflect the data points more clearly. FedScale can achieve 100 concurrent clients participated in each round out of 10000 total clients (orange point), while Flower framework can achieve 1000 concurrent clients out of a total 1 million clients (green point). The plot shows that Flower can achieve both higher concurrent participated client and larger client pool compared with other experiments existing the the recent research papers. Appendix~\ref{appendix:track} gives details of the papers considered.}
    \label{fig:maxclients}
    \vspace{-0.6cm}
\end{figure}

Federated Learning (FL)~\cite{DBLP:conf/aistats/McMahanMRHA17} is an emerging area of research in the machine learning community which aims to enable distributed edge devices (or users) to collaboratively \emph{train} a shared prediction model while keeping their personal data private. At a high level, this is achieved by repeating three basic steps: i) local parameters update to a shared prediction model on each edge device, ii) sending the local parameter updates to a central server for aggregation, and iii) receiving the aggregated model back for the next round of local updates.

% Systems challenges

From a systems perspective, a major bottleneck to FL research is the paucity of frameworks that support scalable execution of FL methods on mobile and edge devices. While several frameworks including Tensorflow Federated~\cite{tensorflowfed,45381} (TFF) and LEAF~\cite{caldas2018leaf} enable experimentation on FL algorithms, they do not provide support for running FL on edge devices. System-related factors such as heterogeneity in the software stack, compute capabilities, and network bandwidth, affect model synchronization and local training. In combination with the choice of the client selection and parameter aggregation algorithms, they can impact the accuracy and training time of models trained in a federated setting. The systems' complexity of FL and the lack of scalable open-source frameworks can lead to a disparity between FL research and production. While closed production-grade systems report client numbers in the thousands or even millions~\cite{hard2019federated}, few research papers use populations of more than 100 clients, as can be seen in Figure~\ref{fig:maxclients}. Even those papers which use more than 100 clients rely on simulations (e.g., using nested loops) rather than actually implementing FL clients on real devices. 

% Flower fills that gap

In this paper, we present \emph{Flower}\footnote{\url{https://flower.dev}}, a novel FL framework, that supports experimentation with both algorithmic and systems-related challenges in FL. Flower offers a stable, language and ML framework-agnostic implementation of the core components of a FL system, and provides higher-level abstractions to enable researchers to experiment and implement new ideas on top of a reliable stack. Moreover, Flower allows for rapid transition of existing ML training pipelines into a FL setup to evaluate their convergence properties and training time in a federated setting. Most importantly, Flower provides support for extending FL implementations to mobile and wireless clients, with heterogeneous compute, memory, and network resources.

As system-level challenges of limited compute, memory, and network bandwidth in mobile devices are not a major bottleneck for powerful cloud servers, Flower provides built-in tools to simulate many of these challenging conditions in a cloud environment and allows for a realistic evaluation of FL algorithms. Finally, Flower is designed with scalability in mind and enables large-cohort research that leverages both a large number of connected clients and a large number of clients training concurrently. We believe that the capability to perform FL at scale will unlock new research opportunities as results obtained in small-scale experiments are not guaranteed to generalize well to large-scale problems. In summary, we make the following contributions:

\squishlist %%\begin{itemize}
  \item We present Flower, a novel FL framework that supports large-cohort training and evaluation, both on real edge devices and on single-node or multi-node compute clusters. This unlocks scalable algorithmic research of real-world system conditions such as limited computational resources which are common for typical FL workloads.
%  scalable algorithmic research and implementation of FL methods on edge devices and servers, including means to simulate real-world system conditions such as
  \item We describe the design principles and implementation details of Flower. In addition to being language- and ML framework-agnostic by design, Flower is also fully extendable and can incorporate emerging %parameter averaging 
  algorithms, training strategies and communication protocols.
  \item Using Flower %as the underlying framework
  , we present experiments that explore both algorithmic and system-level aspects of FL on five machine learning workloads with up to 15 million clients. Our results quantify the impact of various system bottlenecks such as client heterogeneity and fluctuating network speeds on FL performance.
  \item \emph{Flower is open-sourced under Apache 2.0 License}
  and adopted by major research organizations in both academia and industry. The community is actively participating in the development and contributes novel baselines, functionality, and algorithms.
\squishend %%\end{itemize}

\section{Background and Related Work}
FL builds on a vast body of prior work and has since been expanded in different directions. McMahan et al.~\yrcite{DBLP:conf/aistats/McMahanMRHA17} introduced the basic federated averaging (FedAvg) algorithm and evaluated it in terms of communication efficiency. There is active work on privacy and robustness improvements for FL: A targeted model poisoning attack using Fashion-MNIST~\cite{xiao2017} (along with possible mitigation strategies) was demonstrated by Bhagoji et al.~\yrcite{DBLP:journals/corr/abs-1811-12470}. Abadi et al.~\yrcite{45428} propose an attempt to translate the idea of differential privacy to deep learning. Secure aggregation~\cite{Bonawitz:2017:PSA:3133956.3133982} is a way to hide model updates from ``honest but curious" attackers. Robustness and fault-tolerance improvements at the optimizer level are commonly studied and demonstrated, e.g., by Zeno~\cite{pmlr-v97-xie19b}. Finally, there is an increasing emphasis on the performance of federated optimization in heterogeneous data and system settings~\cite{smith2017federated, li2018federated, li2019fair}.

The optimization of distributed training with and without federated concepts has been covered from many angles \cite{Dean:2012:LSD:2999134.2999271,DBLP:journals/corr/abs-1807-11205,DBLP:journals/corr/abs-1810-11787,DBLP:journals/corr/abs-1802-05799,Dryden:2016:CQD:3018874.3018875}. Bonawitz et al.~\yrcite{47976} detail the system design of a large-scale Google-internal FL system. TFF~\cite{tensorflowfed}, PySyft~\cite{DBLP:journals/corr/abs-1811-04017}, and LEAF~\cite{caldas2018leaf} propose open source frameworks which are primarily used for simulations that run a small number of \textit{homogeneous} clients. Flower unifies both perspectives by being open source and suitable for exploratory research, with scalability to expand into settings involving a large number of \textit{heterogeneous} clients. Most of the mentioned approaches have in common that they implement their own systems to obtain the described results. The main intention of Flower is to provide a framework which would (a) allow to perform similar research using a common framework and (b) enable to run those experiments on a large number of \textit{heterogeneous} devices.

\vspace{-0.3cm}
\section{Flower Overview}
% \todo{TODO Basically the same as before, but emphasise scalability, system heterogeneity and privacy.}
% \todo{Research to production. In centralized machine learning, Frameworks such as TensorFlow and PyTorch led to an}

Flower is a novel end-to-end federated learning framework that enables a more seamless transition from experimental research in simulation to system research on a large cohort of real edge devices. Flower offers individual strength in both areas (viz. simulation and real world devices); and offers the ability for experimental implementations to migrate between the two extremes as needed during exploration and development. In this section, we describe use cases that motivate our perspective, design goals, resulting framework architecture, and comparison to other frameworks.

\subsection{Use Cases}
The identified gap between FL research practice and industry reports from proprietary large-scale systems (Figure~\ref{fig:maxclients}) is, at least in part, related a number of use cases that are not well-supported by the current FL ecosystem. The following sections show how Flower enables those use cases.
% We present use cases that are challenging to implement for researchers and engineers given alternative frameworks, and outline the role of Flower in overcoming those challenges.

% \textbf{Efficient resource utilization.} Given a fixed computational budget (e.g., single-machine/multi-GPU), researchers need efficient FL frameworks with little overhead and resource-aware execution. Efficient resource utilization enables FL algorithms to be tested at reasonable scale (client pool size) and speed (wall-clock execution time).

\textbf{Scale experiments to large cohorts.} Experiments need to scale to both a large client pool size and a large number of clients training concurrently to better understand how well methods generalize. A researcher needs to be able launch large-scale FL evaluations of their algorithms and design using reasonable levels of compute (e.g., single-machine/a multi-GPU rack), and have results at this scale have acceptable speed (wall-clock execution time).

\textbf{Experiment on heterogeneous devices.} Heterogeneous client environments are the norm for FL. Researchers need ways to both simulate heterogeneity and to execute FL on real edge devices to quantify the effects of system heterogeneity. Measurements about the performance of client performance should be able to be easily collected, and deploying heterogeneous experiments is painless. %Such experiments need to be 

\textbf{Transition from simulation to real devices.} New methods are often conceived in simulated environments. To understand their applicability to real-world scenarios, frameworks need to support seamless transition between simulation and on-device execution. Shifting from simulation to real devices, mixing simulated and real devices, and selecting certain elements to have varying levels of realism (e.g., compute or network) should be easy.

\textbf{Multi-framework workloads.} Diverse client environments naturally motivate the usage of different ML frameworks, so FL frameworks should be able to integrate updates coming from clients using varying ML frameworks in the same workload. Examples range from situations where clients use two different training frameworks (pytorch and tensorflow) to more complex situations where clients have their own device- and OS-specific training algorithm.

\begin{table}[t]
\caption{Excerpt of built-in FL algorithms available in Flower. New algorithms can be implemented using the \emph{Strategy} interface.}
% \vskip 0.15in
\centering
\small
\begin{tabular}{@{}cl@{}}
\toprule
\textbf{Strategy} & \multicolumn{1}{c}{\textbf{Description}} \\ \midrule
FedAvg & \begin{tabular}[c]{@{}l@{}}Vanilla Federated Averaging ~\cite{DBLP:conf/aistats/McMahanMRHA17}\end{tabular} \\ \midrule
\begin{tabular}[c]{@{}c@{}}Fault \\ Tolerant \\ FedAvg\end{tabular} & \begin{tabular}[c]{@{}l@{}}A variant of FedAvg that can tolerate faulty client\\ conditions such as client disconnections or laggards.\end{tabular} \\ \midrule
FedProx & \begin{tabular}[c]{@{}l@{}}Implementation of the algorithm proposed by\\ Li et al.~\yrcite{li2020federated} to extend FL to heterogenous \\ network conditions.\end{tabular} \\ \midrule
QFedAvg & \begin{tabular}[c]{@{}l@{}}Implementation of the algorithm proposed by \\ Li et al.~\yrcite{li2019fair} to encourage fairness in FL.\end{tabular} \\ \midrule
%\begin{tabular}[c]{@{}c@{}}FedFS \\ (unpub.)\end{tabular} & \begin{tabular}[c]{@{}l@{}}A new strategy for FL in scenarios of \\ heterogeneous client computational capabilities. \\ This  strategy identifies fast and slow clients\\ and intelligently  schedules FL rounds across \\ them to minimize the total convergence time.\end{tabular} \\ \midrule
FedOptim & \begin{tabular}[c]{@{}l@{}} A family of server-side optimizations that \\ include FedAdagrad, FedYogi, and FedAdam \\ as described in
Reddi et al.~\yrcite{reddi2021adaptive}.\end{tabular} \\ \bottomrule
\end{tabular}
% \vspace{-0.8cm}
\label{tab:strategies}
\end{table}

\subsection{Design Goals}
\label{subsec.goals}
The given uses cases identify a gap in the existing FL ecosystem that results in research that does not necessarily reflect real-world FL scenarios. To adress the ecosystem gap, we defined a set of independent design goals for Flower:

%\begin{itemize}%[leftmargin=*,topsep=0pt]
\textbf{Scalable:} Given that real-world FL would encounter a large number of clients, Flower should scale to a large number of concurrent clients to foster research on a realistic scale.

\textbf{Client-agnostic:} Given the heterogeneous environment on mobile clients, Flower should be interoperable with different programming languages, operating systems, and hardware.% settings.

\textbf{Communication-agnostic:} Given the heterogeneous connectivity settings, Flower should allow different serialization and communication approaches.

\textbf{Privacy-agnostic:} Different FL settings (cross-devic, cross-silo) have different privacy requirements (secure aggregation, differential privacy). Flower should support common approaches whilst not be prescriptive about their usage.

\textbf{Flexible:} Given the rate of change in FL and the velocity of the general ML ecosystem, Flower should be flexible to enable both experimental research and adoption of recently proposed approaches with low engineering overhead.
%\end{itemize}

A framework architecture with those properties will increase both realism and scale in FL research and provide a smooth transition from research in simulation to large-cohort research on real edge devices. The next section describes how the Flower framework architecture supports those goals.  

\subsection{Core Framework Architecture}
FL can be described as an interplay between global and local computations. Global computations are executed on the server side and responsible for orchestrating the learning process over a set of available clients. Local computations are executed on individual clients and have access to actual data used for training or evaluation of model parameters.

\begin{figure}[t]
    \small
    \centering
    \includegraphics[width=0.4\textwidth]{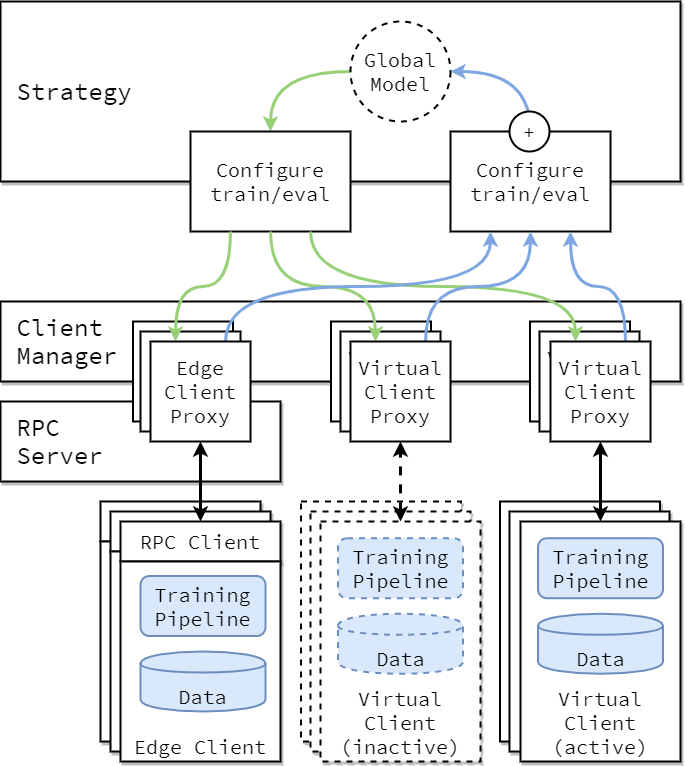}
    \vspace{-0.3cm}
    \captionsetup{font=small,labelfont=bf}
    \caption{Flower core framework architecture with both Edge Client Engine and Virtual Client Engine. Edge clients live on real edge devices and communicate with the server over RPC. Virtual clients on the other hand consume close to zero resources when inactive and only load model and data into memory when the client is being selected for training or evaluation.}% for training or evaluation.\\}
    \vspace{-0.3cm}
    \label{fig:framework-architecture}
\end{figure}

The architecture of the Flower core framework reflects that perspective and enables researchers to experiment with building blocks, both on the global and on the local level. Global logic for client selection, configuration, parameter update aggregation, and federated or centralized model evaluation can be expressed through the \emph{Strategy} abstraction. An implementation of the \emph{Strategy} abstraction represents a single FL algorithm and Flower provides tested reference implementations of popular FL algorithms such as FedAvg~\cite{DBLP:conf/aistats/McMahanMRHA17} or FedYogi~\cite{reddi2021adaptive} (summarized in table~\ref{tab:strategies}). Local logic on the other hand is mainly concerned with model training and evaluation on local data partitions. Flower acknowledges the breadth and diversity of existing ML pipelines and offers ML framework-agnostic ways to federate these, either on the \emph{Flower Protocol} level or using the high-level \emph{Client} abstraction. Figure \ref{fig:framework-architecture} illustrates those components.

The Flower core framework implements the necessary infrastructure to run these workloads at scale. On the server side, there are three major components involved: the \emph{ClientManager}, the FL loop, and a (user customizable) \emph{Strategy}. Server components sample clients from the \emph{ClientManager}, which manages a set of \emph{ClientProxy} objects, each representing a single client connected to the server. They are responsible for sending and receiving \emph{Flower Protocol} messages to and from the actual client. The FL loop is at the heart of the FL process: it orchestrates the entire learning process. % and ensures that progress is made.
It does not, however, make decisions about \emph{how} to proceed, as those decisions are delegated to the currently configured \emph{Strategy} implementation.

In summary, the FL loop asks the \emph{Strategy} to configure the next round of FL, sends those configurations to the affected clients, receives the resulting client updates (or failures) from the clients, and delegates result aggregation to the \emph{Strategy}. It takes the same approach for both federated training and federated evaluation, with the added capability of server-side evaluation (again, via the \emph{Strategy}). The client side is %(architecturally)
simpler in the sense that it only waits for messages from the server. It then reacts to the messages received by calling user-provided training and evaluation functions.

A distinctive property of this architecture is that the server is unaware of the nature of connected clients, which allows to train models across heterogeneous client platforms and implementations, including workloads comprised of clients connected through different communication stacks. The framework manages underlying complexities such as connection handling, client life cycle, timeouts, and error handling in an for the researcher.% without prescribing a particular federation algorithm.

\subsection{Virtual Client Engine}
Built into Flower is the Virtual Client Engine (VCE): a tool that enables the virtualization of Flower Clients to maximise utilization of the available hardware. Given a pool of clients, their respective compute and memory budgets (e.g. number of CPUs, VRAM requirements) and, the FL-specific hyperparameters (e.g. number of clients per round), the VCE launches Flower Clients in a resource-aware manner. The VCE will schedule, instantiate and run the Flower Clients in a transparent way to the user and the Flower Server. This property greatly simplifies parallelization of jobs, ensuring the available hardware is not underutilised and, enables porting the same FL experiment to a wide varying of setups without reconfiguration: a desktop machine, a single GPU rack or multi-node GPU cluster. The VCE therefore becomes a key module inside the Flower framework enabling running large scale FL workloads with minimal overhead in a scalable manner.

%Since the VCE is aware of the amount of compute and memory resources available for the FL experiment at hand it can 

%Clients with a resource (e.g. CPU and/or VRAM) budget assigned to them.  

\subsection{Edge Client Engine}
Flower is designed to be open source, extendable and, framework and device agnostic. Some devices suitable for lightweight FL workloads such as Raspberry Pi or NVIDIA Jetson require minimal or no special configuration. These Python-enabled embedded devices can readily be used as Flower Clients. On the other hand, commodity devices such as smartphones require a more strict, limited and sometimes proprietary software stack to run ML workloads. To circumvent this limitation, Flower provides a low-level integration %possibility 
by directly handling \emph{Flower Protocol} messages on the client.

% \hl{Flower is open and device/framework agnostic but running on certain devices requires extra components. We have those for Android, iOS. Maybe we can borrow something from On-device-ML workshop paper from Akhil}

% \begin{figure}[t]
%     \small
%     \centering
%     \includegraphics[width=0.40\textwidth]{mlsys2022style/diagrams/flower-core-framework-architecture-HD Edge Client Engine.png}
%     \vspace{-0.3cm}
%     \captionsetup{font=small,labelfont=bf}
%     \caption{Flower core framework architecture with Edge Client Engine. Each edge client lives on a real edge device and communicates with the server via RPC.\\}
%     \vspace{-0.6cm}
%     \label{fig:framework-architecture}
% \end{figure}

\subsection{Secure Aggregation}\label{sec:sa}
In FL the server does not have direct access to a client's data. To further protect clients' local data, Flower provides implementation of both SecAgg \cite{Bonawitz:2017:PSA:3133956.3133982} and SecAgg$+$ \cite{bell2020secure} protocols for a semi-honest threat model. The Flower secure aggregation implementation satisfies five goals: usability, flexibility, compatibility, reliability and efficiency. The execution of secure aggregation protocols is independent of any special hardware and ML framework, robust against client dropouts, and has lower theoretical overhead for both communication and computation than other traditional multi-party computation secure aggregation protocol, which will be shown in \ref{sec:secure_exp}.

\subsection{FL Framework Comparison}
%TODO Emphasise scalability, system heterogeneity (ML framework, language, hardware), privacy.

%\hl{TODO More head-to-head (compare modules). Include figure here. }

%TODO FedScale information. only provide PyTorch recipe. has server-side definitions. achieve client availability (e.g., device drop-out or rejoining).

We compare Flower to other FL toolkits, namely TFF \cite{tensorflowfed}, Syft~\cite{DBLP:journals/corr/abs-1811-04017}, FedScale~\cite{lai2021fedscale} and LEAF~\cite{caldas2018leaf}. Table~\ref{tab.features} provides an overview, with a more detailed description of those properties following thereafter.

\begin{table}[t]
    \small
    \captionsetup{font=small,labelfont=bf}
    \caption{Comparison of different FL frameworks.}
    \vskip 0.15in
    \centering
    \setlength\tabcolsep{3pt}
    \scalebox{0.9}{
    \begin{tabular}{|c|c c c c c|}
        \toprule
         & \textbf{TFF} & \textbf{Syft} & \textbf{FedScale} & \textbf{LEAF} & \textbf{Flower} \\
        \midrule
        \midrule
         Single-node simulation & $\surd$ & $\surd$ & $\surd$ & $\surd$ & $\surd$ \\
         \hline
         Multi-node execution & * & $\surd$ & ($\surd$)*** & & $\surd$ \\
         \hline
         Scalability & * & & ** & & $\surd$ \\
         \hline
         Heterogeneous clients & & ($\surd$)*** & ** &  & $\surd$ \\
         \hline
         ML framework-agnostic &  & **** & **** & & $\surd$ \\
         \hline
         Communication-agnostic &  &  & & & $\surd$ \\
         \hline
         Language-agnostic &  &  & & & $\surd$ \\
         \hline
         Baselines &  & & $\surd$ & $\surd$ & * \\
        \bottomrule
    \end{tabular}
    }\\
    \vspace{1mm}
    Labels: \textbf{*} Planned / \textbf{**} Only simulated \\ \textbf{***} Only Python-based /  \textbf{****} Only PyTorch and/or TF/Keras 
    \vspace{-0.3cm}
    \label{tab.features}
\end{table}

%\begin{itemize}%[leftmargin=*,topsep=0pt]
\textbf{Single-node simulation} enables simulation of FL systems on a single machine to investigate workload performance without the need for a multi-machine system. Supported by all frameworks.
  
\textbf{Multi-node execution} requires network communication between server and clients on different machines. Multi-machine execution is currently supported by Syft and Flower. FedScale supports multi-machine simulation (but not real deployment), TFF plans multi-machine deployments.

\textbf{Scalability} is important to derive experimental results that generalize to large cohorts. Single-machine simulation is limited because workloads including a large number of clients often exhibit vastly different properties. TFF and LEAF are, at the time of writing, constrained to single-machine simulations. FedScale can simulate clients on multiple machines, but only scales to ~100 concurrent clients. Syft is able to communicate over the network, but only by connecting to data holding clients that act as servers themselves, which limits scalability. In Flower, data-holding clients connect to the server which allows workloads to scale to millions of clients, including scenarios that require full control over when connections are being opened and closed. Flower also includes a virtual client engine for large-scale multi-node simulations.

\textbf{Heterogeneous clients} refers to the ability to run workloads comprised of clients running on different platforms using different languages, all in the same workload. FL targeting edge devices will clearly have to assume pools of clients of many different types (e.g., phone, tablet, embedded). Flower supports such heterogeneous client pools through its language-agnostic and \emph{communication-agnostic} client-side integration points. It is the only framework in our comparison that does so, with TFF and Syft expecting a framework-provided client runtime, whereas FedScale and LEAF focus on Python-based simulations.

\textbf{ML framework-agnostic} toolkits allow researchers and users to leverage their previous investments in existing ML frameworks by providing universal integration points. This is a unique property of Flower: the ML framework landscape is evolving quickly (e.g., JAX~\cite{jax2018github}, PyTorch Lightning~\cite{pytorchlightning2019github}) and therefore the user should choose which framework to use for their local training pipelines. TFF is tightly coupled with TensorFlow and experimentally supports JAX, LEAF also has a dependency on TensorFlow, and Syft provides hooks for PyTorch and Keras, but does not integrate with arbitrary tools.

\textbf{Language-agnostic} describes the capability to implement clients in a variety of languages, a property especially important for research on mobile and emerging embedded platforms. These platforms often do not support Python, but rely on specific languages (Java on Android, Swift on iOS) for idiomatic development, or native C++ for resource constrained embedded devices. Flower achieves a fully language-agnostic interface by offering protocol-level integration. Other frameworks are based on Python, with some of them indicating a plan to support Android and iOS (but not embedded platforms) in the future.

\textbf{Baselines} allow the comparison of existing methods with new FL algorithms. Having existing implementations at ones disposal can greatly accelerate research progress. LEAF and FedScale come with a number of benchmarks built-in with different datasets. TFF provides libraries for constructing baselines with some datasets. Flower currently implements a number of FL methods in the context of popular ML benchmarks, e.g., a federated training of CIFAR-10~\cite{CIFAR-10} image classification, and has initial port of LEAF datasets such as FEMNIST and Shakespeare~\cite{caldas2018leaf}.
%\end{itemize}

\section{Implementation}
% TODO Shorten to one page, include SecAgg. and VCE
Flower has an extensive implementation of FL averaging algorithms, a robust communication stack, and various examples of deploying Flower on real and simulated clients. Due to space constraints, we only focus on some of the implementation details in this section and refer the reader to the Flower GitHub repository for more details. 

% This section introduces some implementation details and their relation to the high-level goals of Flower.

\label{sec:implementation}
\textbf{Communication stack.} FL requires stable and efficient communication between clients and server.
The Flower communication protocol is currently implemented on top of bi-directional gRPC~\cite{grpc} streams.
gRPC defines the types of messages exchanged and uses compilers to then generate efficient implementations for different languages such as Python, Java, or C++. A major reason for choosing gRPC was its efficient binary serialization format, which is especially important on low-bandwidth mobile connections. Bi-directional streaming allows for the exchange of multiple message without the overhead incurred by re-establishing a connection for every request/response pair.

\textbf{Serialization.} Independent of communication stack, Flower clients receive instructions (messages) as raw byte arrays (either via the network or throught other means, for example, inter-process communication), deserialize the instruction, and execute the instruction (e.g., training on local data). The results are then serialized and communicated back to the server. Note that a client communicates with the server through language-independent messages and can thus be implemented in a variety of programming languages, a key property to enable real on-device execution. The user-accessible byte array abstraction makes Flower uniquely serialization-agnostic and enables users to experiment with custom serialization methods, for example, gradient compression or encryption.

\textbf{Alternative communication stacks.} Even though the current implementation uses gRPC, there is no inherent reliance on it. The internal Flower server architecture uses modular abstractions such that components that are not tied to gRPC are unaware of it. This enables the server to support user-provided RPC frameworks and orchestrate workloads across heterogeneous clients, with some connected through gRPC, and others through other RPC frameworks.

\textbf{ClientProxy.} The abstraction that enables communication-agnostic execution is called \emph{ClientProxy}. Each ClientProxy object registered with the ClientManager represents a single client that is available to the server for training or evaluation. Clients which are offline do not have an associated ClientProxy object. All server-side logic (client configuration, receiving results from clients) is built against the ClientProxy abstraction.

One key design decision that makes Flower so flexible is that ClientProxy is an abstract interface, not an implementation. There are different implementations of the ClientProxy interface, for example, GrpcClientProxy. Each implementation encapsulates details on how to communicate with the actual client, for example, to send messages to an actual edge device using gRPC.

\textbf{Virtual Client Engine (VCE).} Resource consumption (CPU, GPU, RAM, VRAM, etc.) is the major bottleneck for large-scale experiments. Even a modestly sized model easily exhausts most systems if kept in memory a million times. The VCE enables large-scale single-machine or multi-machine experiments by executing workloads in a resource-aware fashion that either increases parallelism for better wall-clock time or to enable large-scale experiments on limited hardware resources. It creates a \emph{ClientProxy} for each client, but defers instantiation of the actual client object (including local model and data) until the resources to execute the client-side task (training, evaluation) become available. This avoids having to keep multiple client-side models and datasets in memory at any given point in time.

VCE builds on the Ray~\cite{moritz2018ray} framework to schedule the execution of client-side tasks. In case of limited resources, Ray can sequence the execution of client-side computations, thus enabling a much larger scale of experiments on common hardware. The capability to perform FL at scale will unlock new research opportunities as results obtained in small-scale experiments often do not generalize well to large-cohort settings.

\section{Framework Evaluation}

In this section we evaluate Flower’s capabilities in supporting both research and implementations of real-world FL workloads. Our evaluation focuses on three main aspects:
%\begin{itemize}[leftmargin=*]
\squishlist
    \item \textbf{Scalability}: We show that Flower can (a) efficiently make use of available resources in single-machine simulations and (b) run experiments with millions of clients whilst sampling thousands in each training.
    \item \textbf{Heterogeneity}: We show that Flower can be deployed in real, heterogeneous devices commonly found in cross-device scenario and how it can be used to measure system statistics.
    \item \textbf{Realism}: We show through a case study how Flower can throw light on the performance of FL under heterogeneous clients with different computational and network capabilities.  
    \item \textbf{Privacy}: Finally, we show how our implementation of Secure Aggregation matches the expected theoretical overhead as expected.
\squishend
%\end{itemize}

\subsection{Large-Scale Experiment}
Federated Learning receives most of its power from its ability to leverage data from millions of users. However, selecting large numbers of clients in each training round does not necessarily translate into faster convergence times. In fact, as observed in  \citep{DBLP:conf/aistats/McMahanMRHA17}, there is usually an empirical threshold for which if we increase the number of participating clients per round beyond that point, convergence will be slower. By allowing experiments to run at mega-scales, with thousands of active clients per round, Flower gives us the opportunity to empirically find such threshold for any task at hand. 

To show this ability, in this series of experiments we use Flower to fine-tune a network on data from 15M users using different numbers of clients per round. More specifically, we fine-tune a Transformer network to correctly predict Amazon book ratings based on text reviews from users.

\begin{figure}[t]
    \small
    \centering
    % \includegraphics[width=0.5\textwidth]{diagrams/framework.pdf}
    % \parjump{}
    %\includegraphics[width=0.4\textwidth]{mlsys2022style/diagrams/fig1_colour.pdf}
    \includegraphics[width=0.45\textwidth]{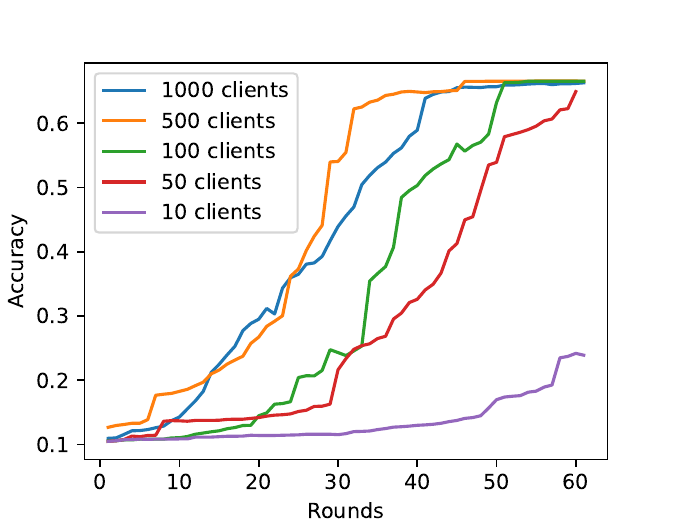}
    \vspace{-0.3cm}
    \captionsetup{font=small,labelfont=bf}
    \caption{Flower scales to even 15M user experiments. Each curve shows successful convergence of the DistilBERT model under varying amounts of clients per round, with the exception of the two smallest client sizes: 50 and 10. 
    }
    %Convergence curves for different numbers of clients per round.}
    %\todo{mention which experiment.(just enough to complete the second line of caption)}}
    
    \label{fig:largescale}
\end{figure}

\textbf{Experimental Setup.} We choose to use Amazon's Book Reviews Dataset \citep{ni2019justifying} which contains over 51M reviews from 15M different users. Each review from a given user contains a textual review of a book along with its given rank (1-5). We fine-tune the classifier of a pre-trained DistilBERT model \cite{sanh2019distilbert} to correctly predict ranks based on textual reviews. For each experiment we fix the number of clients being sampled in each round (from 10 to 1000) and aggregate models using \texttt{FedAvg}. We test the aggregated model after each round on a fixed set of 1M clients. Convergence curves are reported in Figure \ref{fig:largescale} all our experiments were run using two NVIDIA V100 GPUs on a 22-cores of an Intel Xeon Gold 6152 (2.10GHz) CPU.

\textbf{Results.} Figure \ref{fig:largescale} shows the expected initial speed-up in convergence when selecting 10 to 500 clients per round in each experiment. However, if we decide to sample 1k clients in each round, we notice an increase in convergence time. Intuitively, this behaviour is caused by clients' data having very different distributions; making it difficult for simple Aggregation Strategies such as \texttt{FedAvg} to find a suitable set of weights.

\subsection{Single Machine Experiments}
One of our strongest claims in this paper is that Flower can be effectively used in Research. For this to be true, Flower needs to be fast at providing reliable results when experimenting new ideas, e.g.\ a new aggregation strategy. 

In this experiment, we provide a head-to-head comparison in term of training times between Flower and the four main FL frameworks, namely FedScale, TFF, FedJax and the original LEAF, when training with different FL setups.

\textbf{Experimental Setup.} We consider all three FL setups proposed by \citep{caldas2018leaf} when training a CNN model to correctly classify characters from the \textit{FEMNIST} dataset. More specifically, we consider the scenarios where the number of clients ($c$) and local epochs per round change ($l$) vary. The total number of rounds and total number of clients are kept constant at 2000 and 179, respectively. To allow for a fair comparison, We run all our experiments using eight cores of an Intel Xeon E5-2680 CPU (2.40GHz) equipped with two NVIDIA RTX2080 GPUs and 20GB of RAM.

\begin{figure}[t]
    \small
    \centering
    \includegraphics[width=0.45\textwidth]{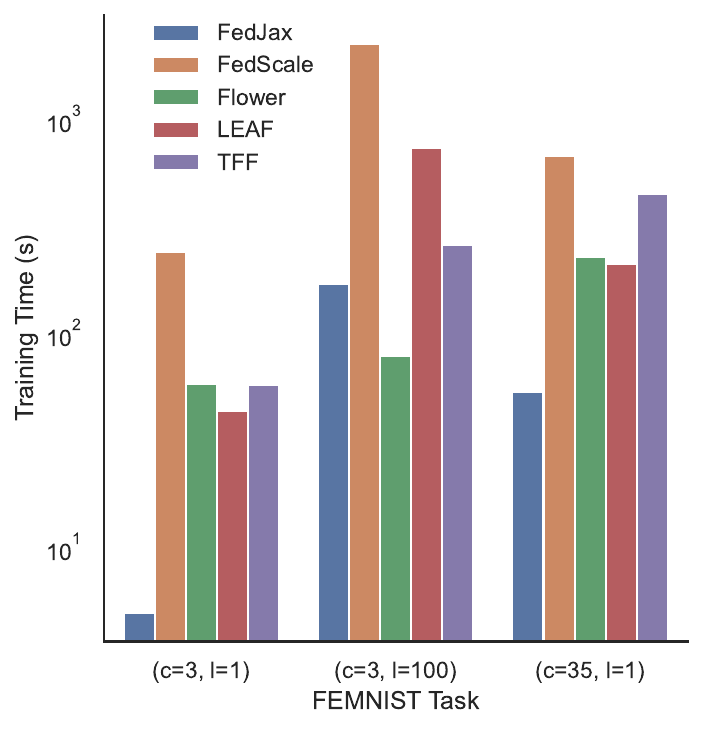}
    \vspace{-0.5cm}
    \captionsetup{font=small,labelfont=bf}
\caption{Training times (log scale in second) comparison of different FEMNIST tasks between different FL frameworks.}
    \label{fig:traing_times}
    \vspace{-0.5cm}
\end{figure}

\textbf{Results.} Figure \ref{fig:traing_times} shows the impact of choosing different FL frameworks for the various tasks. On our first task, when training using three clients per round ($c=3$) for one local epoch ($l=1$), FedJax finishes training first (05:18), LEAF finishes second (44:39) followed by TFF (58:29) and Flower (59:19). In this simple case, the overhead of having a multi-task system, like the Virtual Client Engine (VCE), causes Flower to sightly under-perform in comparison to loop-based simulators, like LEAF.

However, the benefits of having a VCE become more evident if we train on more realistic scenarios. When increasing the number of clients per round to 35 while keeping the single local epoch, we notice that Flower (230:18) is still among the fastest frameworks. Since the number of local epochs is still one, most of the overhead comes from loading data and models into memory rather than performing real training, hence the similarity those LEAF and Flower. 

The VCE allows us to specify the amount of GPU memory we want to associate with each client, this allows for more efficient data and model loading of different clients on the same GPU, making the overall training considerably faster. In fact, when we substantially increase the amount of work performed by each client to 100 local epochs, while fixing the number of active client to 3, we see a significant saving in training time. In this task Flower outperforms all other. It completes the task in just about 80 minutes, while the second best performing framework (FedJax) takes over twice as long (over 173 minutes).  

It is also important to acknowledge the two extreme training times we see in this experiment. FedJax seems to be very efficient when training on few (1) local epochs; however, in scenarios where communication-efficiency is key and larger number of local epochs are required, FedJax performance slightly degrades.   
FedScale, on the other hands, consistently showed high training times across all training scenarios. We believe this apparent inefficiency to be associated with network overheads that are usually unnecessary in a single-computer simulation.

%When experimenting with new ideas for FL, e.g. developing a novel aggregation strategy, it is important to be able to train
%Here we present a head-to-head comparison between different frameworks when training on LEAF datasets on a single machine.

\subsection{Flower enables FL evaluation on real devices}
Flower can assist researchers in quantifying the system costs associated with running FL on real devices and to identify bottlenecks in real-world federated training. In this section, we present the results of deploying Flower on six types of heterogeneous real-world mobile and embedded devices, including Java-based Android smartphones and Python-based Nvidia Jetson series devices and Raspberry Pi.

\begin{figure}[b]
    \centering
    \includegraphics[width=1\linewidth]{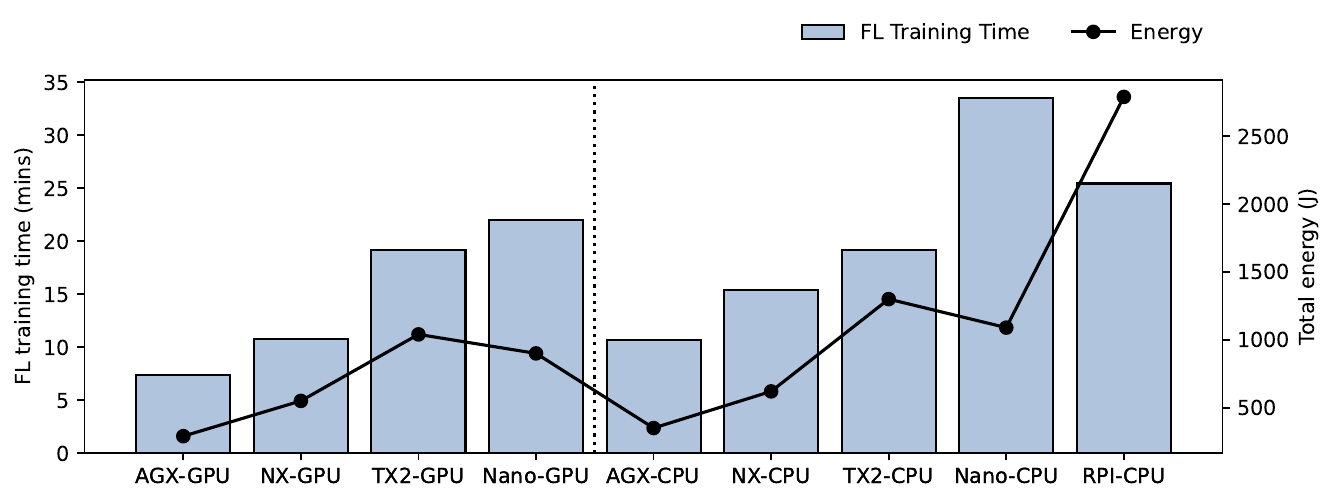}
    \captionsetup{font=small,labelfont=bf}
    \vspace{-0.5cm}
    \caption{Flower enables quantifying the system performance of FL on mobile and embedded devices. Here we report the training times and energy consumption associated with running FL on CPUs and GPUs of various embedded devices.}
    \label{fig:flwr_hetero}
    \vspace{-0.5cm}
\end{figure}

% \begin{table}[t]
%     \small
%     \caption{Flower clients on AWS Device Farm.}
%     \vskip 0.05in
%     \centering
%     \begin{tabular}{|c|c|c|}
%         \toprule
%         \textbf{Device Name} & \textbf{Type} & \textbf{OS Version} \\
%         \midrule
%         Google Pixel 4 & Phone & 10 \\
%         Google Pixel 3 & Phone & 10 \\
%         Google Pixel 2 & Phone & 9 \\
%         Samsung Galaxy Tab S6 & Tablet & 9 \\
%         Samsung Galaxy Tab S4 & Tablet & 8.1.0\\
%         \bottomrule
%     \end{tabular}
%     \vspace{-0.6cm}
%     \label{tab.devices}
% \end{table}

\textbf{Experiment Setup.} We run the Flower server configured with the \texttt{FedAvg} strategy and host it on a cloud virtual machine. Python-based Flower clients are implemented for Nvidia Jetson series devices (Jetson Nano, TX2, NX, AGX) and Raspberry Pi, and trained using TensorFlow as the ML framework on each client.  On the other hand, Android smartphones currently do not have extensive on-device training support with TensorFlow or PyTorch. To counter this issue, we leverage TensorFlow Lite to implement Flower clients on Android smartphones in Java. While TFLite is primarily designed for on-device inference, we leverage its capabilities to do on-device model personalization to implement a FL client application~\cite{tflite}. The source code for both implementations is available in the Flower repository.

% Nvidia TX2 edge devices support full-fledged PyTorch as the ML framework -- this means we could successfully port existing PyTorch training pipelines to implement FL clients on them. 

% Finally, to scale our experiments to a reasonably large number of mobile clients with different OS versions, we deploy Flower on the Amazon AWS Device Farm~\cite{device_farm} that enables deploying applications on real mobile devices accessed through AWS. 

% \begin{table}[t]
% \small
% \captionsetup{font=small,labelfont=bf}
% \caption{Performance metrics with Nvidia Jetson TX2 clients as we vary the number of local epochs. We use the CIFAR10 dataset and train a ResNet-18 model on it. Number of clients $C$ is set to 10 and the model is trained for 40 rounds.}
% \vskip 0.1in
% \centering
% \begin{tabular}{@{}c|ccc@{}}
% \toprule
% \textbf{\begin{tabular}[c]{@{}c@{}}Local\\ Epochs (E)\end{tabular}} & \multicolumn{1}{l}{\textbf{Accuracy}} & \textbf{\begin{tabular}[c]{@{}c@{}}Convergence\\ Time (mins)\end{tabular}} & \textbf{\begin{tabular}[c]{@{}c@{}}Energy\\ Consumed (kJ)\end{tabular}}\\ \midrule
% 1 & 0.48 & 17.63 & 10.21 \\
% 5 & 0.64 & 36.83 & 50.54 \\
% 10 & 0.67 & 80.32 & 100.95 \\ \bottomrule
% \end{tabular}
% \label{tab:jetson}
% \end{table}

\begin{figure}[t]
    \centering
    \includegraphics[width=0.9\linewidth]{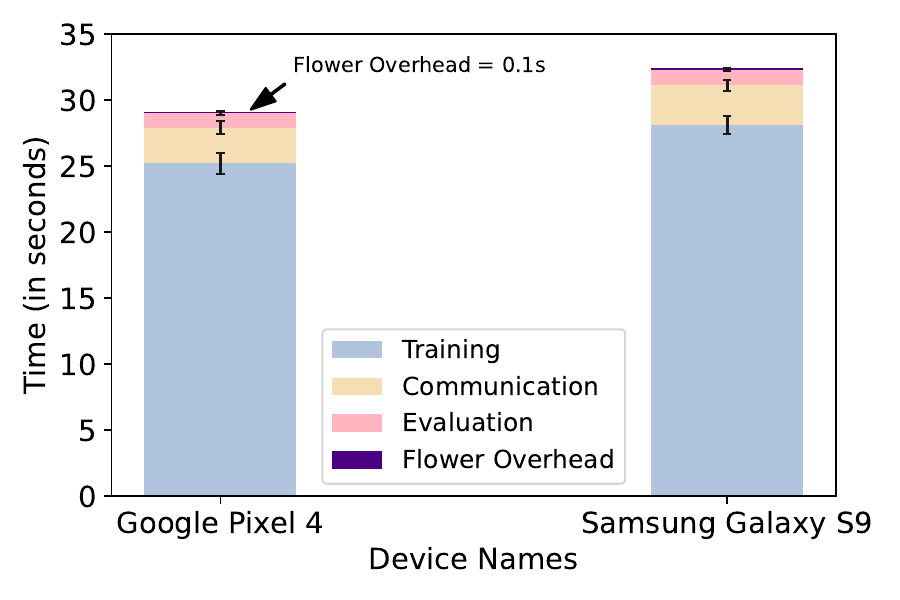}
    \captionsetup{font=small,labelfont=bf}
    \vspace{-0.5cm}
    \caption{Flower enables fine-grained profiling of FL performance on real devices. The framework overhead is $<$100ms per round.}
    \label{fig:flwr_overhead}
   % \vspace{-0.3cm}
\end{figure}

\textbf{Results.} Figure~\ref{fig:flwr_hetero} shows the system metrics associated with training a DeepConvLSTM~\cite{deepconvlstm} model for a human activity recognition task on Python-enabled Jetson and Raspberry Pi devices. We used the RealWorld dataset~\cite{sztyler2016onbody} consisting of time-series data from accelerometer and gyroscope sensors on mobile devices, and partitioned it across 10 mobile clients. The first takeaway from our experiments in that we could deploy Flower clients on these heterogeneous devices, without requiring any modifications in the client-side Flower code. The only consideration was to ensure that a compatible ML framework (e.g., TensorFlow) is installed on each client. Secondly, we show in Figure~\ref{fig:flwr_hetero} how FL researchers can deploy and quantify the training time and energy consumption of FL on various heterogeneous devices and processors. Here, the FL training time is aggregated over 40 rounds, and includes the time taken to perform local 10 local epochs of SGD on the client, communicating model parameters between the server and the client, and updating the global model on the server. By comparing the relative energy consumption and training times across various devices, FL researchers can devise more informed client selection policies that can tradeoff between FL convergence time and overall energy consumption. For instance, choosing Jetson Nano-CPU based FL clients over Raspberry Pi clients may increase FL convergence time by 10 minutes, however it reduces the overall energy consumption by almost 60\%.

Next, we illustrate how Flower can enable fine-grained profiling of FL on real devices. We deploy Flower on 10 Android clients to train a model with 2 convolutional layers and 3 fully-connected layers~\cite{android_cnn} on the CIFAR-10 dataset. TensorFlow Lite is used as the training ML framework on the devices. We measure the time taken for various FL operations, such as local SGD training, communication between the server and client, local evaluation on the client, and the overhead due to the Flower framework. The overhead includes converting model gradients to GRPC-compatible buffers and vice-versa, to enable communication between Java FL clients and a Python FL server. In Figure~\ref{fig:flwr_overhead}, we report the mean latency of various FL operations over 40 rounds on two types of Android devices: Google Pixel 4 and Samsung Galaxy S9. We observe that on both devices, local training remains the most time-consuming operation, and that the total system overhead of the Flower framework is less than 100ms per round.

\subsection{Realism in Federated Learning}
Flower facilitates the deployment of FL on real-world devices. While this property is beneficial for production-grade systems, can it also assist researchers in developing better federated optimization algorithms? In this section, we study two realistic scenarios of FL deployment.

\begin{table}[]
\small
\caption{Effect of computational heterogeneity on FL training times. Using Flower, we can compute a hardware-specific cutoff $\tau$ (in minutes) for each processor, and find a balance between FL accuracy and training time. $\tau=0$ indicates no cutoff time.}
\centering
\begin{tabular}{@{}cccll@{}}
\toprule
\multicolumn{1}{l}{\textbf{}} & \multicolumn{1}{l}{\textbf{GPU}} & \textbf{\begin{tabular}[c]{@{}c@{}}CPU \\ ($\tau$ = 0)\end{tabular}} & \textbf{\begin{tabular}[c]{@{}l@{}}CPU \\ ($\tau$ = 2.23)\end{tabular}} & \textbf{\begin{tabular}[c]{@{}l@{}}CPU \\ ($\tau$ = 1.99)\end{tabular}} \\ \midrule
Accuracy & 0.67 & 0.67 & 0.66 & 0.63 \\
\begin{tabular}[c]{@{}c@{}}Training \\ time (mins)\end{tabular} & 80.32 & \begin{tabular}[c]{@{}c@{}}102\\ (1.27$\times$)\end{tabular} & \begin{tabular}[c]{@{}l@{}}89.15\\ (1.11$\times$)\end{tabular} & \begin{tabular}[c]{@{}l@{}}80.34\\ (1.0$\times$)\end{tabular} \\ \bottomrule
\end{tabular}
\label{tab:compute_hetero}
\end{table}

\textbf{Computational Heterogeneity across Clients.} In real-world, FL clients will have vastly different computational capabilities. While newer smartphones are now equipped with mobile GPUs, other phones or wearable devices may have a much less powerful processor. How does this computational heterogeneity impact FL?

For this experiment, we use a Nvidia Jetson TX2 as the client device, which has a Pascal GPU and six CPU cores. We train a ResNet18 model on the CIFAR-10 dataset in a federated setting with 10 total Jetson TX2 clients and 40 rounds of training. In Table~\ref{tab:compute_hetero}, we observe that if Jetson TX2 CPU clients are used for federated training (local epochs $E$$=$$10$), the FL process would take $1.27\times$ more time to converge as compared to training on Jetson TX2 GPU clients. 

Once we obtain this quantification of computational heterogeneity using Flower, we can design better federated optimization algorithms. As an example, we implemented a modified version of FedAvg where each client device is assigned a cutoff time ($\tau$) after which it must send its model parameters to the server, irrespective of whether it has finished its local epochs or not. This strategy has parallels with the FedProx algorithm \cite{li2018federated} which also accepts partial results from clients. However, the key advantage of Flower's on-device training capabilities is that we can accurately  measure and assign a \emph{realistic} processor-specific cutoff time for each client. For example, we measure that on average it takes 1.99 minutes to complete a FL round on the TX2 GPU. We then set the same time as a cutoff for CPU clients ($\tau = 1.99$ mins) as shown in Table~\ref{tab:compute_hetero}. This ensures that we can obtain faster convergence even in the presence of CPU clients, at the expense of a 4\% accuracy drop. With $\tau = 2.23$, a better balance between accuracy and convergence time could be obtained for CPU clients.

% Certainly, the computational power of each client will impact the time taken to perform local parameter updates on the shared model, which in turn would influence the total time for FL. 

% A researcher interested in doing such an experiment to evaluate the effect of compute heterogeneity on convergence time could simply implement and run Flower Clients on the platform of interest and execute them with the Flower framework to obtain the total FL training time and convergence properties. 

% In Table~\ref{tab:compute_hetero}, we show a result of synchronous training a ResNet50 \cite{DBLP:journals/corr/HeZR016} model on the CIFAR-10 dataset with 10 clients. First, we only use GPU-enabled machines as the FL clients and obtain the total training time for 60 rounds of FL -- for example, with $E=5$, it takes around 270 minutes to collaboratively train the model on GPU (Nvidia V100) machines. Next, we add just one CPU-only machine to the client pool, i.e., the model is now trained with 9 GPU-enabled clients and 1 CPU-only client. We observe that the training time increases to 970 minutes (3.5x) (for $E=5$) due to the computational bottleneck of the CPU-only machine. While this finding is not surprising, it can certainly assist a researcher or a developer to appropriately schedule their FL workloads.   

\textbf{Heterogeneity in Network Speeds.} An important consideration for any FL system is to choose a set of participating clients in each training round. In the real-world, clients are distributed across the world and vary in their download and upload speeds. Hence, it is critical for any FL system to study how client selection can impact the overall FL training time. We now present an experiment with 40 clients collaborating to train a 4-layer deep CNN model for the FashionMNIST dataset. More details about the dataset and network architecture are presented in the Appendix.

\begin{figure}[b]
    \centering
    \includegraphics[width=0.8\linewidth]{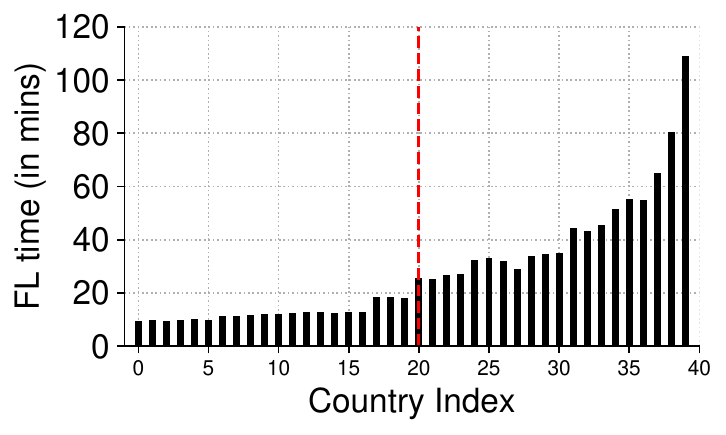}
    \vspace{-0.3cm}
    \caption{Effect of network heterogeneity in clients on FL training time. Using this quantification, we designed a new client sampling strategy called FedFS (detailed in the Appendix).}
  % \vspace{-0.3cm}
    \label{fig:communication}
\end{figure}

Using Flower, we instantiate 40 clients on a cloud platform and fix the download and upload speeds for each client using the \textsc{Wondershaper} library. Each client is representative of a country and its download and upload speed is set based on a recent market survey of 4G and 5G speeds in different countries \cite{opensignal}. 

The x-axis of Figure~\ref{fig:communication} shows countries arranged in descending order of their network speeds: country indices 1-20 represent the top 20 countries based on their network speeds (mean download speed = 40.1Mbps), and indices 21-40 are the bottom 20 countries (mean download speed = 6.76Mbps). We observe that if all clients have the network speeds corresponding to Country 1 (Canada), the FL training finishes in 8.9 mins. As we include slower clients in FL, the training time gradually increases, with a major jump around index = 17. On the other extreme, for client speeds corresponding to Country 40 (Iraq), the FL training takes 108 minutes. 

There are two key takeaways from this experiment: a) Using Flower, we can profile the training time of any FL algorithm under scenarios of network heterogeneity, b) we can leverage these insights to design sophisticated client sampling techniques.
%For example, in the first FL round, we could compute the training times of each client, and use them to cluster the clients. For the subsequent FL rounds, we can randomly choose a cluster and sample clients from it for training.
For example, during subsequent rounds of federated learning, we could monitor the number of samples each client was able to process during a given time window and increase the selection probability of slow clients to balance the contributions of fast and slow clients to the global model.
The FedFS strategy detailed in the appendix works on this general idea, and reduces the convergence time of FL by up to 30\% over the FedAvg random sampling approach.
%which randomly samples clients each round.

\subsection{Secure Aggregation Overheads}\label{sec:secure_exp}
%Although a central server cannot directly access clients' data in a FL setting, it is possible for a malicious client to inspect clients' trained models and perform inference attacks on them to gain information on their data.  Therefore, it is important that we can aggregate trained models in a secure way such that a malicious server cannot inspect trained models individually, especially when the data being trained on are very sensitive. 
\begin{figure}
    \centering
     \begin{subfigure}[b]{0.3\textwidth}   
        \centering
        \begin{tikzpicture}[scale=0.60]
\begin{axis}[align =center,
    title={},
    xlabel={Model vector size},
    ylabel={CPU running time (s)},
    xlabel near ticks,
    ylabel near ticks,
    xmin=0, xmax=600000,
    ymin=0, ymax=300,
    xtick={0,100000,200000,300000,400000,500000},
    ytick={0,50,100, 150,200,250,300},
    legend pos=north west,
    ymajorgrids=true,
    grid style=dashed,
]

\addplot[
    color=blue,
    mark=*
    ]
    coordinates {
    (100000,98.58514578)(200000,107.1396354)(300000, 117.7148892)(400000, 124.9209483)(500000, 134.721084)
    };

\addplot[
    color=red,
    mark=*
    ]
    coordinates {
    (100000, 140.9449628)(200000,172.0023658)(300000,200.6615239)(400000, 229.6977153)(500000, 259.2806949)
    };
    \legend{0\% dropout,5\% dropout}
    
\end{axis}
\end{tikzpicture}
        % \caption{Running time of server with increasing vector size}
        \label{fig:secagg_time}
    \end{subfigure} 
    % \begin{subfigure}[b]{0.2\textwidth}   
    %     \centering
    %     \input{mlsys2022style/sections/communication}
    %     \caption{Total data transfer  with increasing vector size}
    %     \label{fig:secagg_data}
    % \end{subfigure}
    % \captionsetup{font=small,labelfont=bf}
    \vspace{-0.1cm}
    \caption{Performance of Secure Aggregation. Running time of server with increasing vector size
        \vspace{-0.2cm}
%\hl{maybe put these two aligned horizontally and place fig3 here (which will fit whne removing shakespeare)}
    }
\end{figure}
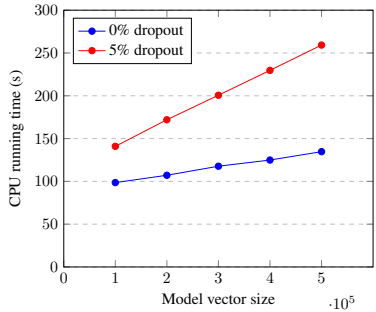

Privacy is one of the cornerstones in FL, which inevitably generates computational overhead during training. In hardware-constrained systems, such as cross-device FL, it is desirable not only to be able to measure such overheads, but also to make sure that security protocols are well implemented and follow the expected protocol described in the original papers. Flower's implementation of Secure Aggregation, named Salvia, is based on the SecAgg \cite{Bonawitz:2017:PSA:3133956.3133982} and SecAgg$+$ \cite{bell2020secure} protocols as described in Section \ref{sec:sa}. To verify that Salvia's behavior matches the expected theoretical complexity, we evaluate its impact on server-side computation and communication overhead with the model vector size and clients dropouts.  

\textbf{Experiment Setup.} The FL simulations run on a Linux system with an Intel Xeon E-2136 CPU ($3.30$GHz), with $256$ GB of RAM. In our simulations, all entries of our local vectors are of size $24$ bits. We ignore communication latency. % in our simulations.
Moreover, all dropouts simulated happen after stage $2$, i.e. Share Keys Stage. This is because this  imposes the most significant overhead as the server not only needs to regenerate dropped-out clients' secrets, but also compute their pairwise masks generated between their neighbours. 

For  our simulations,  the $n$ and $t$ parameters of the $t$-out-of-$n$ secret-sharing scheme are set to 51 and 26, respectively. These parameters are  chosen to reference SecAgg$+$'s proven correctness and  security guarantees, where we can tolerate up to $5\%$  dropouts and $5\%$ corrupted clients with correctness holding with probability $1-2^{-20}$ and security holding with probability $1-2^{-40}$.

\textbf{Results. } Fixing the number of sampled clients to $100$, we plotted CPU running times 
%and total data transfer of the server measured 
through aggregating a vector of size $100$k entries to aggregating one of size $500$k entries in
Figure 8%figure~\ref{fig:secagg_time}% 
%and \ref{fig:secagg_data}, respectively
. We also measured how the performance would change after client dropouts by repeating the same experiments with a $5\%$ client dropout. 

%We observe that 
Both the running times and total data transfer of the server increase linearly with the model vector size as the operations involving model vectors are linear to the vectors' sizes, e.g.\ generating masks, sending vectors. We also note the server's running time increases when there are $5\%$ clients dropping out, as the server has to perform extra computation to calculate all $k$ pairwise masks for each client dropped. %out in the Unmask Vectors Stage.
Lastly, we observe that the total data transferred of the server remains unchanged with client dropouts as each client only communicates with the server plus exactly $k$ neighbors, regardless of the total number of clients and dropouts. We conclude that all our experimental data matches the expected %computation and communication 
complexities of SecAgg and SecAgg$+$.% protocols.

\section{Conclusion}
We have presented Flower -- a novel framework that is specifically designed to advance FL research by enabling heterogeneous FL workloads at scale. Although Flower is broadly useful across a range of FL settings, we believe that it will be a true \textit{game-changer} for reducing the disparity between FL research and real-world FL systems. Through the provided abstractions and components, researchers can federated existing ML workloads (regardless of the ML framework used) and transition these workloads from large-scale simulation to execution on heterogeneous edge devices.
We further evaluate the capabilities of Flower in experiments that target both scale and systems heterogeneity by scaling FL up to 15M clients,  providing head-to-head comparison between different FL frameworks for single-computer experiments,  measuring FL energy consumption on a cluster of Nvidia Jetson TX2 devices, optimizing convergence time under limited bandwidth, and illustrating a deployment of Flower on a range of Android mobile devices in the AWS Device Farm.
Flower is open-sourced under Apache 2.0 License and we look forward to more community contributions to it.

\appendix
%Details on current implementation such as available aggregation methods, etc.
\bibliography{references}
\bibliographystyle{mlsys2022}

%%%%%%%%%%%%%%%%%%%%%%%%%%%%%%%%%%%%%%%%%%%%%%%%%%%%%%%%%%%%%%%%%%%%%%%%%%%%%%%
%%%%%%%%%%%%%%%%%%%%%%%%%%%%%%%%%%%%%%%%%%%%%%%%%%%%%%%%%%%%%%%%%%%%%%%%%%%%%%%
% SUPPLEMENTAL CONTENT AS APPENDIX AFTER REFERENCES
%%%%%%%%%%%%%%%%%%%%%%%%%%%%%%%%%%%%%%%%%%%%%%%%%%%%%%%%%%%%%%%%%%%%%%%%%%%%%%%
%%%%%%%%%%%%%%%%%%%%%%%%%%%%%%%%%%%%%%%%%%%%%%%%%%%%%%%%%%%%%%%%%%%%%%%%%%%%%%%
\appendix
\section{Appendix}
%\section{Please add supplemental material as appendix here}
%\input{sections/amazon_books}
%\input{sections/femnist}
\subsection{Survey on papers}\label{appendix:track}
\begin{figure}[t]
    \small
    \centering
    % \includegraphics[width=0.5\textwidth]{diagrams/framework.pdf}
    % \parjump{}
    %\includegraphics[width=0.4\textwidth]{mlsys2022style/diagrams/fig1_colour.pdf}
    \includegraphics[width=0.45\textwidth]{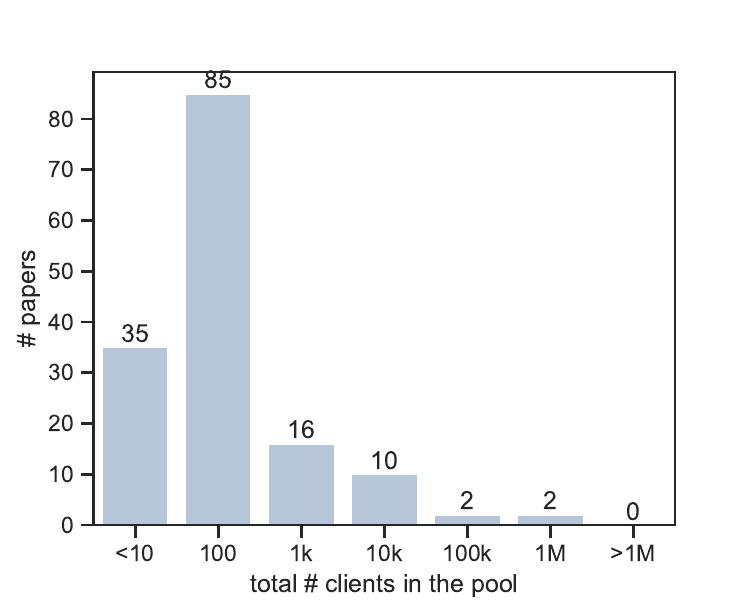}
    \vspace{-0.3cm}
    \captionsetup{font=small,labelfont=bf}
    \caption{Histograms of the number of total FL clients used in FL research papers in the last two years. A vast majority of papers only use up to 100 clients.}
    \label{fig:hist}
\end{figure}

From a systems perspective, a major bottleneck to FL research is the paucity of frameworks that support scalable execution of FL methods on mobile and edge devices. Fig. \ref{fig:hist} shows the histograms of total number of clients in the FL pools in research papers. The research papers is gathered from Google Scholar that is related to federated learning from last 2 years which consists of total $150$ papers in the survey. We excluded papers that are using the framework not available to reproduced the results. As we can see from the histogram, the majority of experiments only use up to $100$ total clients, which usually on datasets such as CIFAR10 and ImageNet. There are only $3$ papers using the dataset with a total clients pool up to $1$ millions, and they are using the Reddit and Sentiment140 dataset from leaf \cite{caldas2018leaf}. 

\subsection{FedFS Algorithm}
\label{appendix:fedfs}

We introduce \emph{Federating: Fast and Slow} (\emph{FedFS}) to overcomes the challenges arising from heterogeneous devices and non-IID data. \emph{FedFS} acknowledges the difference in compute capabilities inherent in networks of mobile devices by combining partial work, importance sampling, and dynamic timeouts to enable clients to contribute equally to the global model.

\textbf{Partial work}. Given a (local) data set of size $m_k$ on client $k$, a batch size of $B$, and the number of local training epochs $E$, FedAvg performs $E \frac{m_k}{B}$ (local) gradient updates $\theta^{k} \leftarrow \theta^{k} - \eta \triangledown \ell (b ; \theta^{k})$ before returning $\theta^{k}$ to the server. The asynchronous setting treats the success of local update computation as binary. If a client succeeds in computing $E \frac{m_k}{B}$ mini-batch updates before reaching a timeout $\Delta$, their weight update is considered by the server, otherwise it is discarded. The server then averages all successful $\theta_{k \in \{ 0, .., K\}}$ updates, weighted by $m_k$, the number of training examples on client $k$.

This is wasteful because a clients' computation might be discarded upon reaching $\Delta$ even if it was close to computing the full $E \frac{m_k}{B}$ gradient updates. We therefore apply the concept of partial work \cite{li2018federated} in which a client submits their locally updated $\theta_k$ upon reaching $\Delta$ along with $c_k$, the number of examples actually involved in computing $\theta_k$, even if $c_k < E \frac{m_k}{B} B$. The server averages by $c_k$, not $m_k$, because $c_k$ can vary over different rounds and devices depending on a number of factors (device speed, concurrent processes, $\Delta$, $m_k$, etc.).

Intuitively, this leads to more graceful performance degradation with smaller values for $\Delta$. Even if $\Delta$ is set to an adversarial value just below the completion time of the fastest client, which would cause \emph{FedAvg} to not consider any update and hence prevent convergence, \emph{FedFS} would still progress by combining $K$ partial updates. More importantly it allows devices which regularly discard their updates because of lacking compute capabilities to have their updates represented in the global model, which would otherwise overfit the data distribution on the subset of faster devices in the population.

% \paragraph{Importance sampling} Partial work enables \emph{FedFS} to leverage $c_k^t$, the amount of work done by client $k$ during round $t$, for client selection during round $t+1$. $c_k < m_k$ suggests that client $k$ is not able to compute $E \frac{m_k}{B}$ gradient updates in less than $\Delta$, so its weight update $\theta_k$ has less of an impact on the global model $\theta$ compared to an update from client $j$ with $c_j = m_j$. \emph{FedFS} acknowledges this difference by alternating between two client selection schemes to perform $r_{f}$ fast rounds and $r_{s}$ slow rounds. During fast rounds clients are selected with a probability proportional to $\frac{1}{k} \frac{c_k}{m_k}$ (normalized over all clients), whereas slow rounds select clients with $1 - \frac{c_k}{m_k}$ to increase the probability of selecting ``slow" clients (i.e., $c_k < m_k$). The distinction between fast and slow rounds allows clients which had low contributions in previous rounds to be selected more frequently during slow rounds, and hence contribute additional updates to the global model. \emph{FedAvg} is a special case of \emph{FedFS}: if all clients are able to compute $c_k = m_k$ every round, then there are no clients to be selected during slow rounds, and \emph{FedFS} performs only fast rounds with client selection probability $\frac{1}{k}$.

\textbf{Importance sampling.} Partial work enables \emph{FedFS} to leverage the observed values for $c_k^r$ (with $r \in \{1, ..., t\}$, the amount of work done by client $k$ during all previous rounds up to the current round $t$) and $E^r m_k$ (with $r \in \{1, ..., t\}$, the amount of work client $k$ was maximally allowed to do during those rounds) for client selection during round $t+1$. $c$ and $m$ can be measured in different ways depending on the use case. In vision, $c_k^t$ could capture the number of image examples processed, whereas in speech $c_k^t$ could measure the accumulated duration of all audio samples used for training on client $k$ during round $t$. $c_k^t < E^t m_k$ suggests that client $k$ was not able to compute $E^t \frac{m_k}{B}$ gradient updates within $\Delta_t$, so its weight update $\theta_k^t$ has less of an impact on the global model $\theta$ compared to an update from client $j$ with $c_j^t = E^t m_j$.
\emph{FedFS} uses importance sampling for client selection to mitigate the effects introduced by this difference in client capabilities.
We define the work contribution $w_k$ of client $k$ as the ratio between the actual work done during previous rounds 
$c_k = \sum_{r=1}^{t} c_k^r$ and the maximum work possible $\hat{c}_{k} = \sum_{r=1}^{t} E^r m_k$. Clients which have never been selected before (and hence have no contribution history) have $w_k=0$. We then sample clients on the selection probability $1 - w_k + \epsilon$ (normalized over all $k \in \{1, ..., K\}$), with $\epsilon$ being the minimum client selection probability. $\epsilon$ is an important hyper-parameter that prevents clients with $c_k^t = E^t m_k$ to be excluded from future rounds. Basing the client selection probability on a clients' previous contributions ($w_k$) allows clients which had low contributions in previous rounds to be selected more frequently, and hence contribute additional updates to the global model. Synchronous \emph{FedAvg} is a special case of \emph{FedFS}: if all clients are able to compute $c_k^t = E^t m_k$ every round, then there will be no difference in $w_k$ and \emph{FedFS} samples amongst all clients with a uniform client selection probability of $\frac{1}{k}$.

\textbf{Alternating timeout.} Gradual failure for clients which are not able to compute $E^t \frac{m_k}{B}$ gradient updates within $\Delta_t$ and client selection based on previous contributions allow \emph{FedFS} to use more aggressive values for $\Delta$. One strategy is to use an alternating schedule for $\Delta$ in which we perform $r_f$ ``fast" rounds with small $\Delta^f$) and $r_s$ ``slow" rounds with larger $\Delta^s$. This allows \emph{FedFS} to be configured for either improved convergence in terms of wall-clock time or better overall performance (e.g., in terms for classification accuracy).
% Our evaluation results \todo{add ref} demonstrate that considerable improvements in terms of wall-clock time convergence can be achieved by alternating between fast and slow rounds.

\textbf{FedFS algorithm.} The full \emph{FedFS} algorithm is given in Algorithm \ref{algo:fedfs}.

\begin{algorithm2e}[t]
  \small
  \SetAlgoLined
  \DontPrintSemicolon
  \Begin(Server $T, C, K, \epsilon, r_f, r_s, \Delta_{max}, E, B,$){
    initialise $\theta_0$\;
    \For{round $t \leftarrow 0, ..., T-1$}{
        $j \leftarrow \max ( \lfloor C \cdot K \rfloor , 1) $ \;
        
        $\mathcal{S}_t \leftarrow$ (sample $j$ distinct indices from $\{ 1, ..., K \}$ with $1 - w_k + \epsilon$) \;
        
        \eIf{fast round ($r_f, r_s$)}{
            $\Delta_t = \Delta^f$ \;
        }{
            $\Delta_t = \Delta^s$ \;
        }
        
        \ForPar{$k \in \mathcal{S}_t$}{
            $\theta_{t+1}^{k}, c_k, m_k \leftarrow$ ClientTraining($k$, $\Delta_t$, $\theta_t$, $E$, $B$, $\Delta_t$) \;
        }
        $c_r \leftarrow \sum_{k \in \mathcal{S}_t}^{} c_k$ \;
        $\theta_{t+1} \leftarrow \sum_{k \in \mathcal{S}_t} \; \frac{c_k}{c_r} \; \theta_{t+1}^{k}$ \;
    }
  }
  \caption{FedFS}
  \label{algo:fedfs}
\end{algorithm2e}

\definecolor{orange}{rgb}{1, 0.49, 0.17}
\definecolor{matgreen}{rgb}{0.09, 0.62, 0.21}

\begin{figure}[t]
\begin{center}
\scalebox{0.75}{

\begin{tikzpicture}
\begin{axis}[
    width=7cm, height=5cm,   % size of the image
    grid = major,
    grid style={dashed, gray!30},
    xmin=1,   % start the diagram at this x-coordinate
    xmax=15,  % end  the diagram at this x-coordinate
    ymin=0,   % start the diagram at this y-coordinate
    ymax=85,  % end  the diagram at this y-coordinate
    axis background/.style={fill=white},
    xtick={1,2,3,4,5,6,7,8,9,10,11,12,13,14,15},
    xticklabels={1,2,3,4,5,,,,,10,,,,,15},
    tick align=outside,
    mark repeat={600},
    ylabel near ticks,
    xlabel near ticks,
    xlabel={Training time (Days)},
    ylabel={Accuracy (\%)},
    legend pos=south east,
    legend columns=1,
    legend style={
                    % the /tikz/ prefix is necessary here...
                    % otherwise, it might end-up with `/pgfplots/column 2`
                    % which is not what we want. compare pgfmanual.pdf
            /tikz/column 2/.style={
                column sep=2pt,
            },
    }
    ]

\addplot[color=orange,line width=1pt] file {diagrams/imagenet_acc1.txt};%\addlegendentry{ASR $\rightarrow$ ASR}
\addlegendentry{Top-1 Accuracy}

\addplot+[mark=none,color=matgreen,line width=1pt] file {diagrams/imagenet_acc5.txt};%\addlegendentry{ASR $\rightarrow$ ASR}
\addlegendentry{Top-5 Accuracy}

%\addplot+[mark=none,color=red,line width=1pt] file %{ressources/liGRU_ep.txt};%\addlegendentry{ASR $\rightarrow$ ASR}
%\addlegendentry{liGRU}

% resultats_DECODA_SANS_IVECTORS_DEV_corpus_RES_RES_RES_FALSE_EFR_2IT.txt 0.7952457142857122
% MEDIANE : 0.794285714285714 A LA POSITION 250.5 SUR 500

\end{axis}
\end{tikzpicture}

}
\vspace{-0.4cm}
\caption{Training time reported in days and accuracies (Top-1 and Top-5) for an ImageNet federated training with Flower.}\label{fig:fl_imagenet}
\vspace{-0.6cm}
\end{center}
\end{figure}
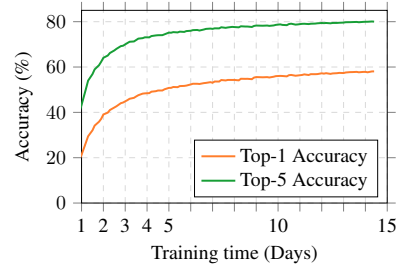

\subsection{Scaling FedAvg to ImageNet-scale datasets} We now demonstrate that Flower can not only scale to a large number of clients, but it can also support training of FL models on web-scale workloads such as ImageNet. To the best of our knowledge, this is the first-ever attempt at training ImageNet in a FL setting.    

\textbf{Experiment Setup}. We use the ILSVRC-2012 ImageNet partitioning \citep{russakovsky2015imagenet} that contains $1.2M$ pictures for training and a subset composed of $50K$ images for testing. We train a ResNet-18 model on this dataset in a federated setting with $50$ clients equipped with four physical CPU cores. To this end, we partition the ImageNet training set into 50 IID partitions and distribute them on each client. During training, we also consider a simple image augmentation scheme based on random horizontal flipping and cropping. 

% ImageNet is expected to demonstrate the scalability of Flower to realistic federated learning scenarios with large datasets. 

% For our experiment, we generate randomly $50$ subsets from the official train set corresponding to $50$ clients following an IID scheme. 

\textbf{Results}. Figure \ref{fig:fl_imagenet} shows the results on the test set of ImageNet obtained by training a ResNet-18 model. It is worth to mention that based on $50$ clients and $3$ local epochs, the training lasted for about $15$ days demonstrating Flower's potential to run long-term and realistic experiments. 

% In our setup, all the clients were located within the same compute node, however Flower could have handled a longer training time by introducing networking latencies to simulate a real-world traffic. 

We measured top-1 and top-5 accuracies of $59.1$\% and $80.4$\% respectively obtained with FL compared to $63$\% and $84$\% for centralised training. First, it is clear from Figure \ref{fig:fl_imagenet} that FL accuracies could have increased a bit further at the cost of a longer training time, certainly reducing the gap with centralised training. Then, the ResNet-18 architecture relies heavily on  batch-normalisation, and it is unclear how the internal statistics of this technique behave in the context of FL, potentially harming the final results. As expected, the scalability of Flower helps with raising and investing new issues related to federated learning. 

For such long-term experiments, one major risk is that client devices may go offline during training, thereby nullifying the training progress. Flower's built-in support for keeping the model states on the server and resuming the federated training from the last saved state in the case of failures came handy for this experiment.

\subsection{Datasets and Network Architectures}

We use the following datasets and network architectures for our experiments. 

\textbf{CIFAR-10} consists of 60,000 images from 10 different object classes. The images are 32 x 32 pixels in size and in RGB format. We use the training and test splits provided by the dataset authors --- 50,000 images are used as training data and remaining 10,000 images are reserved for testing. 

\textbf{Fashion-MNIST} consists of images of fashion items (60,000 training, 10,000 test) with 10 classes such as trousers or pullovers. The images are 28 x 28 pixels in size and in grayscale format. We use a 2-layer CNN followed by 2 fully-connected layers for training a model on this dataset.  

\textbf{ImageNet}. We use the ILSVRC-2012 ImageNet  \citep{russakovsky2015imagenet} containing $1.2M$ images for training and $50K$ images for testing. A ResNet-18 model is used for federated training this dataset. 

%
%Put anything that you might normally include after the references as an appendix here, {\it not in a separate supplementary file}. Upload your final camera-ready as a single pdf, including all appendices.

%%%%%%%%%%%%%%%%%%%%%%%%%%%%%%%%%%%%%%%%%%%%%%%%%%%%%%%%%%%%%%%%%%%%%%%%%%%%%%%
%%%%%%%%%%%%%%%%%%%%%%%%%%%%%%%%%%%%%%%%%%%%%%%%%%%%%%%%%%%%%%%%%%%%%%%%%%%%%%%

\end{document}